# A GA Based approach for selection of local features for recognition of handwritten Bangla numerals

N. Das[*], S. Basu[*], P. K. Saha[+], R. Sarkar[*], M. Kundu[*], M. Nasipuri

*Computer Sc. & Engg. Dept., Jadavpur University,
Kolkata-700032, India.
[+]Electrical and Computer Engineering, The University of Iowa
Iowa City, IA 52242,USA

**Abstract:**

Soft computing approaches are mainly designed to address the real world ill-defined, imprecisely formulated problems, combining different kind of novel models of computation, such as neural networks, genetic algorithms (GAs. Handwritten digit recognition is a typical example of one such problem. In the current work we have developed a two-pass approach where the first pass classifier performs a coarse classification, based on some global features of the input pattern by restricting the possibility of classification decisions within a group of classes, smaller than the number of classes considered initially. In the second pass, the group specific classifiers concentrate on the features extracted from the selected local regions, and refine the earlier decision by combining the local and the global features for selecting the true class of the input pattern from the group of candidate classes selected in the first pass. To optimize the selection of local regions a GA based approach has been developed here. The maximum recognition performance on *Bangla* digit samples as achieved on the test set, during the first pass of the two pass approach is 93.35%. After combining the results of the two stage classifiers, an overall success rate of 95.25% is achieved.

## 1. Introduction:

Soft computing approaches are mainly designed to address the real world ill-defined, imprecisely formulated problems, combining different kind of novel models of computation, such as Bayesian Classifiers, neural networks, fuzzy sets and systems, and Genetic Algorithms (GAs), requiring huge *computation* [1]. Handwritten digit recognition is a typical example of one such problem. To identify handwritten digits of varying shapes and sizes, caused by different handwriting styles of different individuals, perceptual power of cognitive capabilities of human beings is required. Due to large varieties of potential applications like extracting data from filled in forms, automatic postal code identification, mail sorting systems, automatic reading of bank cheques etc, handwritten digit recognition is considered as an important problem. *Bangla*, the second most popular language in India and also the national language of Bangladesh, is the fifth most popular language in the world. But despite its importance and popularity, evidences of research on OCR of handwritten *Bangla* characters, as observed in the literature, are a few in numbers.

## 1.1 The previous work

In the recent years it has been noticed that researchers prefer two stage multiple classifier combination techniques [2-14] for character recognition to improve the recognition performance both in time and accuracy. Tulyakov et al. [3] justified the efficiency of different classifier combination methods to increase the performance of pattern recognition applications.

In the work of Huang and Suen [4], a technique called Behavior-Knowledge Space (BKS) method is presented to aggregate the decisions obtained from individual classifiers and derive the best final decisions from the statistical point of view. The technique is applied for recognition of handwritten Roman digits.

Among the other significant contributions in this field, a multiple classifier combination technique was proposed by Ho et al. [5] for recognition of degraded machine printed characters. Xu et.al. [6] had designed a classifier combination technique for improving recognition accuracy of different classifiers for recognition of handwritten Roman numerals. In the work of Govindaraju et.al. [7], a fuzzy classifier combination rule is designed for improvement of recognition accuracy on handwritten Roman numerals. In one of our earlier works [8], recognition performance for Optical Character Recognition (OCR) of handwritten Bangla digits are improved through combination of classification decisions, obtained from two Multi Layer Perceptron (MLP) classifiers, by application of Dempster Shafer (DS) technique.

The work described in [9] involves a two stage hierarchical approach for OCR of handwritten Bangla alphabetic characters, in which multiple experts are employed in the second stage, i.e. after coarse classification, for final classification of a pattern of an unknown class. Classification decisions, in the second stage, are mainly based on the consensus among multiple classifiers but, to reach the consensus, sample confidences of the experts are considered instead of majority voting method.

In [10], Oh et.al. proposed a new approach for combination of multiple features in handwriting recognition problem. In the said scheme two different sets of feature vectors are designed, where one feature-set is designed to be used by all the classes and in the other set, class-dependent feature sets are designed for patterns of each class. A combination algorithm is finally developed to combine the said feature sets for classification of patterns using a neural network classifier.

In the work of Bhattacharya et. al., [11], a two-stage approach is adopted to classify 50 handwritten Basic characters and 10 numeric digits of Bangla script. In this approach also a coarse or a group based coarse classification of an unknown pattern in first stage is followed by a finer classification in the second stage. Based on the similarity of shapes, 57 pattern classes are identified for final classification. These pattern classes are clustered into 11 groups for coarse classification. An MLP based classifier is employed in the first stage to decide about the group of an unknown pattern. In the second stage, the pattern is subjected to another MLP based classifier, specific to its group, for final classification. In another work, Bhattacharya et. al., [12] had proposed a similar two stage approach for recognition of 50 Basic characters of handwritten Bangla script. Chain-code histogram features are used in both the cases for classification through MLP based classifiers.

In a work of the S.Basu et al. , a two pass approach [14] was introduced for pattern classification and was applied for OCR of handwritten Bangla digits. In the first pass of this technique, an unknown digit pattern is coarsely classified into a group of digit classes on the basis of some pre selected global features. Then in the second pass, the classification decision thus obtained in the first pass is refined to determine the within group class membership of the unknown digit pattern on the basis of some group specific local features. Performances of the system are tested with a limited training set and a test set of 300 and 200 randomly selected sample patterns respectively.

**1.2. Motivation behind the current work**

The major objective of the two pass approach presented here is to make the search for a true pattern class in the decision space more focused or guided towards the goal on the basis of the coarse classification decision produced in the first pass. The objective is so set that improved recognition performance is achieved by optimum selection of local features in the second pass of the two pass approach. To do so, a genetic algorithm (GA) based optimization technique is employed for selection of informative sub images from the digit patterns for extraction of aforementioned local features. The classifiers used for the two pass approach are expected to be simple as they work with global and local features separately in stages. Compared to the conventional classifier combination approach, the two pass approach, presented here, does not involve selection of a computationally intensive combiner function to work. This has been the key motivation behind the current work.

## 2 Present work:

In the current two-pass approach, the first classifier performs a *coarse classification*, based on some *global features* of the input pattern by restricting the possibility of classification decisions within a group of classes, smaller than the number of classes considered initially. In the second pass, the classifiers concentrate on the features extracted from the *optimized local regions*, and refine the earlier decision by combining the local and the global features for selecting the true class of the input pattern from the group of candidate classes selected in the first pass. The groups are formed with the pattern classes with high rates of mutual misclassification observed from the *confusion matrix* formed by applying the first pass classifier on the *training data*.To optimize the local regions a GA based approach, as discussed earlier has been developed in the current work. Actions of the two classifiers, which work in sequence, thus constitute two passes of the classification process. The two pass approach creates a scope for correct classification of certain patterns, which are misclassified by the top choices of the first pass classifier. True classes for such patterns may be identified by the group specific classifiers, designed for the second pass of the two pass approach.

To establish authenticity of the proposed two pass approach, it is applied on handwritten *Bangla* digit recognition problem, which is a realistic benchmark problem of pattern recognition. To recognize the *Bangla* handwritten digits, a Multi Layer Perceptron (MLP), i.e., a feed forward model of the artificial neural network with learning and generalization abilities, is used as a pattern classifier. Typical digit patterns of first ten natural numbers of *Bangla* are shown in Fig. 1.

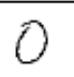

Fig. 1. The decimal digit set of Bengali script

In the first pass of the two pass approach, a 24 element feature set, comprising of global shadow features, is extracted from the overall digit images. From the confusion matrix, obtained during the training of the MLP classifier in the first pass, we have found two groups of mutually misclassifying pattern classes. The labels of the digit patterns of these two mutually misclassifying groups are observed as {'1','9'} and {'0', '3', '4', '5', '6'}. Our objective is to identify the regions of dissimilarity from each such group of digit patterns and subsequently extract the region-specific features, called the local features. These features are extracted by dividing any digit image by a number of overlapping regions, often called as windows. Finding a selective combination of some of these windows may be encoded as an optimization problem. This motivates the use of genetic algorithm (GA), i.e., an intelligent optimization algorithm free from chances of sticking to local minima, in the current work.

### 2.1. Features for Coarse Classification

As mentioned earlier, we have used 24 modified shadow features [16] as global feature set for coarse classification of an unknown pattern in first pass.. Before extraction of these features from a digit image, it is binarized and scaled to a size of 32x32 pixels.

### 2.1 Formation of Groups for Coarse Classification

A coarse classification decision in the two pass approach confines the search for the true class of an unknown pattern within a group of candidate pattern classes smaller than the original one, i.e. the group of all possible pattern classes. This decision is made on the basis of the top choice of the first pass classifier.

By observing the *confusion matrix* developed with classification decisions of the first pass classifier on the *training data*, groups of pattern classes with high rates of mutual misclassifications are determined. The confusion matrices are shown in Fig 2. Once these predefined groups are available, a coarse classification decision is made by identifying the group of an unknown pattern by the top choice class of the first pass classifier.

The groups of pattern classes as formed from the confusion matrices are given below.
Group#1 = {1,9}
Group#2 = {0,3,4,5,6}

|   | 0 | 1 | 2 | 3 | 4 | 5 | 6 | 7 | 8 | 9 |
|---|---|---|---|---|---|---|---|---|---|---|
| 0 : | 394 | 2 | 0 | 1 | 0 | 3 | 0 | 0 | 0 | 0 |
| 1 : | 0 | 371 | 1 | 0 | 0 | 0 | 0 | 0 | 0 | 13 |
| 2 : | 0 | 1 | 398 | 0 | 0 | 1 | 2 | 1 | 0 | 2 |
| 3 : | 1 | 0 | 0 | 393 | 0 | 0 | 7 | 2 | 0 | 0 |
| 4 : | 0 | 1 | 0 | 0 | 395 | 2 | 1 | 2 | 0 | 2 |
| 5 : | 5 | 0 | 0 | 1 | 3 | 390 | 3 | 2 | 0 | 2 |
| 6 : | 0 | 2 | 0 | 3 | 1 | 2 | 384 | 0 | 0 | 1 |
| 7 : | 0 | 0 | 1 | 1 | 1 | 1 | 2 | 391 | 0 | 2 |
| 8 : | 0 | 0 | 0 | 0 | 0 | 0 | 0 | 1 | 400 | 0 |
| 9 : | 0 | 23 | 0 | 1 | 0 | 1 | 1 | 1 | 0 | 378 |

**Fig. 2.** The confusion matrices produced after coarse classification of training samples of handwritten Bangla digits are shown

If the classification decision produced by the first pass classifier does not belong to any such group, the decision is final and does not require refinement in the two pass approach. Depending on the group selected by the first pass classifier, a group of specific classifier is invoked in the second pass for producing the final classifier decision

**2.3. Feature Sets for Various Groups**

Each classifier, which may be invoked in the second pass depending on the group selected in the first pass, is designed to focus the process of pattern classification on certain specific portions of the pattern of an unknown class. It is so because patterns of different classes within a group differ significantly within those portions. Since each such group is formed with pattern classes having high rates of mutual misclassification, the other portions, where the patterns of different classes within the group do not differ much, need not be required for finer classification. Our objective is to identify these regions of dissimilarities from each such group of digit patterns and subsequently extract region-specific features, called the local features. To find the optimized regions, often referred to as windows, for our experiment we have used genetic algorithm (GA) , in the current work. For systematic selection of windows, each digit image of size 32x32 Pixels is divided through 9 equal size overlapping windows of dimension (h/2 x w/2), where h and w are the height and the width of each digit pattern. The 9 overlapping windows are created for the present work with the coordinates of the top left corners as follows-  {(0, 0) , (w/2, h/2)},{ (0,h/2), (w/2, h)},{ (w/4,0), (3*w/4, h/2)},{ (w/2,0), (w, h/2)},{ (w/2, h/2), (w, h)},{ (w/2,h/4), (w, 3*h/4) },{ (w/4, 0), (3*w/4, h/2)}, { (w/4, h/2), (3*w/4,h)},{ (w/4, h/4), (3*w/4,3*h/4)}. Within the windows, longest run features [15] are computed in four directions, viz. row wise, column wise and along the directions of two major diagonals. They are used as region specific local features.

**2.3.1 Selection of sub images using genetic algorithm (GA)**

Selection of a combination of overlapping windows on a digit image, which when used for extraction of longest run features produces an optimal recognition performance, is performed here using GA. In doing this, each candidate solution or chromosome is encoded with a 9 bit binary string, in which each bit corresponds to one of the 9 overlapping windows on a digit image. Each bit in the chromosome has a value 1 if the corresponding window is selected for feature extraction. Otherwise, it has a value 0. Initially a population of candidate solutions is created by randomly generating 20 chromosomes for GA. After every iteration of GA, the population size is kept fixed to 20 chromosomes here.

The fitness of each chromosome of the population thus created is measured by obtaining the recognition rate of an MLP classifier on the test samples, from which longest run features are extracted by considering a combination of overlapping windows, encoded by the chromosome with global shadow features. Prior to that, the MLP classifier is to be trained with the same features extracted from the training samples. For computing fitness values of all the chromosomes of a population this process is to be repeated for each chromosome of the population.

After a population is evaluated on the basis of the fitness function, the stopping criterion of GA is to be tested. In this work, the stopping criterion is reached either after 20 generations have passed or the average fitness value of the current population is greater than or equal to 98% of the maximum fitness value obtained so far.

Failing to reach the stopping criterion, a new population of more promising chromosomes, also called a generation, is to be reproduced from the existing population, in GA. For this, 20 chromosomes are selected here from the existing population using roulette wheel, weighted proportionally with the fitness values of the chromosomes of the existing population. Once a new population of 20 chromosomes is created in this way, 80% of the chromosomes are selected pair wise randomly from the population for performing crossover operation over them. The crossover point is selected at the middle of each chromosome here.

Once Crossover operation is completed, the current population is left with 16 chromosomes, undergoing crossover, plus 4 chromosomes, selected by roulette wheel method. Half of the chromosomes of this population is again randomly selected for mutation. The exact bit of a chromosome to be mutated is also selected randomly. In this way, a generation of population is reproduced from the old one, which is to be again evaluated to check if the stopping criterion is met. GA is repeatedly executed in this way until the stopping criterion is reached.

Thus the optimum combination of regions for each group of digit patterns are finally selected for extraction of local region-specific features during the second pass of the two pass approach. Based on the selected optimized features with global shadow features, the MLP based second pass classifiers for each group finally recognizes the true label of the unknown digit pattern.

**2.3.2. Features for Group#1 and Group#2**

*Bangla* digit patterns of different classes belonging to Group#1 differ in several rectangular sub-images of the minimum bounding box covering each digit image. The genetic algorithm finds that selective sub-images and optimizes the recognition performance for the Group. For Group#1 the regions with coordinates {(w/2,0) (w,h/2), (0,h/2) (w/2,h) , (w/2, h/4) (w, 3*h/4) } and for Group#2 the sub-images with coordinates {(w/2,h/2) (w,h), (w/2, h/4) (w, 3*h/4) } are finally selected. These are illustrated in Fig 3.

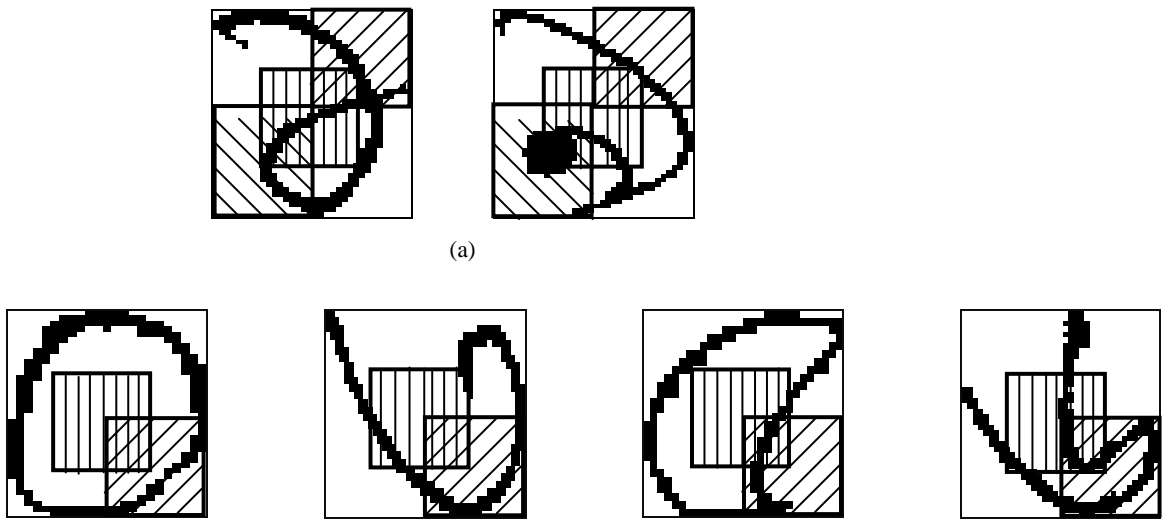

(a)

(b)

**Fig. 3(a).** Shaded regions show the subimages selected by the GA, where patterns of different classes of Group#1$_1$ differ most.

**(b).** Shaded regions show the subimages selected by the GA, where patterns of different classes of Group#2 differ most.

## 3. Experimental results and discussion

To conduct experiments with the technique, described so far, a database of 6000 samples of handwritten *Bangla* digits, each of size 32x32 pixels is prepared by [1]collecting handwritten samples from people of different age groups. Before extraction of features from these sample images, each image is binarized through thresholding. The training set and the test set, used for this work, consist of 4000 samples and 2000 samples of digit patterns respectively.

In the first pass, a coarse classification is done on the basis of[4] the 24 global shadow features. The maximum recognition performance on *Bangla* digit samples as achieved on the test set, during the first pass of the two pass approach is 93.35%. During the second pass, GA is used on each of these two groups to find the optimal regions for extraction of local features. With varying hidden layer neurons the maximum success rates achieved for the two groups are 91% and 96.10% respectively. After combining the results of the two stage classifiers, an overall success rate of 95.25% is achieved. This is shown in Table 1. In all these classification rejection rate is considered as zero.

**Table 1:** Maximum recognition performances of the classifiers during the current two-pass approach on handwritten *Bangla* digit samples.

|  | Coarse (pass -1) | Group1 {'1','9'} | Group2 {'0', '3', '4', '5', '6'}. | Coarse + Fine (Pass2) | Increase in Recognition Performance |
|---|---|---|---|---|---|
| Recognition performance | 93.35% | 91% | 96.10% | 95.25% | 1.9% |
| Description of MLP classifier designed with | 24-35-10 | 36-20-10 | 32-40-10 | ----------- | ----------- |

Compared to the recognition performance of the single pass classifier(93.35%), designed for *Bangla* digit patterns, the recognition performances of the two pass classifier(95.25%), designed for the same, is found to be improved by 1.9%.

As observed from this experiment, the current two pass approach significantly improves the recognition performance of a benchmark pattern recognition problem, in comparison to the conventional single stage approaches. The technique employs an intelligent selection of pattern sub-images for design of the local features in the second pass of the said two pass approach. The coarse and finer classification decisions are combined without the use of complex combiner function. The designed technique may be useful in improving recognition accuracy for patterns belonging to large number of target classes with an optimum tradeoff between the recognition accuracy and computational efficiency.

## Acknowledgements:

Authors are thankful to the "Center for Microprocessor Application for Training Education and Research", "Project on Storage Retrieval and Understanding of Video for Multimedia" of Computer Science & Engineering Department, Jadavpur University, for providing infrastructure facilities during progress of the work.
## Acknowledgements:

Authors are thankful to the "Center for Microprocessor Application for Training Education and Research", "Project on Storage Retrieval and Understanding of Video for Multimedia" of Computer Science & Engineering Department, Jadavpur University, for providing infrastructure facilities during progress of the work.

## Authors' Biography

**Nibaran Das** received his B.Tech degree in Computer Science and Technology from Kalyani Govt. Engineering College under Kalyani University, in 2003. He received his M.C.S.E degree from Jadavpur University, in 2005. He joined J.U. as a lecturer in 2006. His areas of current research interest are OCR of handwritten text, Bengali fonts, biometrics and image processing. He has been an editor of Bengali monthly magazine "Computer Jagat" since 2005.

**Subhadip Basu** received his B.E. degree in Computer Science and Engineering from Kuvempu University, Karnataka, India, in 1999. He received his Ph.D. (Engg.) degree thereafter from Jadavpur University (J.U.) in 2006. He joined J.U. as a senior lecturer in 2006. His areas of current research interest are OCR of handwritten text, gesture recognition, real-time image processing.

**Punam Kumar Saha** received his BCSE, MCSE degrees from Jadavpur University, in 1987, 1989 respectively. He received his Ph.D. (Engg.) degree thereafter from Indian Statistical Institute in 2006. He is a Associate Professor of department of Electrical and Computer Engineering, The University of Iowa, His current research interest include Tensor scale-based image analysis;

digital topology and geometry; fuzzy distance transform, virtual bone biopsy; fuzzy connectedness-based object segmentation; etc

**Ram Sarkar** received his B.Tech degree in Computer Science and Engineering from University of Calcutta, in 2003. He received his M.C.S.E degree from Jadavpur University, in 2005. He joined J.U. as a lecturer in 2008. His areas of current research interest are document image processing, line extraction and segmentation of handwritten text images.

**Mahantapas Kundu** received his B.E.E, M.E.Tel.E and Ph.D. (Engg.) degrees from Jadavpur University, in 1983, 1985 and 1995, respectively. Prof. Kundu has been a faculty member of J.U since 1988. His areas of current research interest include pattern recognition, image processing, multimedia database, and artificial intelligence.

**Mita Nasipuri** received her B.E.Tel.E., M.E.Tel.E., and Ph.D. (Engg.) degrees from Jadavpur University, in 1979, 1981 and 1990, respectively. Prof. Nasipuri has been a faculty member of J.U since 1987. Her current research interest includes image processing, pattern recognition, and multimedia systems. She is a senior member of the IEEE, U.S.A., Fellow of I.E (India) and W.B.A.S.T, Kolkata, India.

**Author Address:** Nibaran Das**,** Computer Sc. & Engineering. Department, Jadavpur University, Kolkata-700032, India , **Phone No:** +913325139162, **Email:** nibaran@cse.jdvu.ac.in